\def\year{2020}\relax
\documentclass[letterpaper]{article} 
\usepackage{aaai20}  
\usepackage{times}  
\usepackage{helvet} 
\usepackage{courier}  
\usepackage[hyphens]{url}  
\usepackage{graphicx} 
\usepackage{graphicx}  
\usepackage{amsmath}
\usepackage{subcaption}
\title{Attending to Mathematical Language with Transformers}
\author{Artit Wangperawong \\ 
artit.wangperawong@usbank.com\\
U.S. Bank\\ 
1095 Avenue of the Americas\\
New York, NY 10036 
}
 
\begin{document}

\maketitle

\begin{abstract}
Mathematical expressions were generated, evaluated and used to train neural network models based on the transformer architecture. The expressions and their targets were analyzed as a character-level sequence transduction task in which the encoder and decoder are built on attention mechanisms. Three models were trained with common hyperparameters to understand and evaluate symbolic variables and expressions in mathematics: (1) the self-attentive and feed-forward transformer without recurrence or convolution, (2) the universal transformer with recurrence, and (3) the adaptive universal transformer with recurrence and adaptive computation time. The models respectively achieved test accuracies as high as 76.1\%, 78.8\% and 84.9\% in evaluating the expressions to match the target values. For the cases inferred incorrectly, the results differed from the targets by only one or two characters. The models notably learned to add, subtract and multiply both positive and negative decimal numbers of variable digits assigned to symbolic variables.
\end{abstract}

\section{Introduction}
Arithmetic and algebra are important mathematical skills that should be acquired by one's adolescence \cite{1carraher2006arithmetic}. Therefore, we should expect that an artificially intelligent agent or system can at least master such problems without predetermined algorithms. Arithmetic involves the study of numbers and the effect on them of operators such as addition ($+$), subtraction ($-$), multiplication ($\times$), and division ($\div$). Algebra at the basic level involves the study of mathematical symbols and the rules governing how such symbols are manipulated. A mathematical expression is a phrase constructed with a finite arrangement of numbers, symbols and operators according to the rules of mathematics. Such rules are typically pre-programmed into computers and execute with ideally perfect accuracy. Here we describe neural network models trained to read mathematical phrases at the character level and evaluate the expressions for a result without any pre-programmed or hard-coded math rules.

\section{Background and Related Work}
Prior studies related to this work have used multilayer perceptrons \cite{2hoshen2016visual}, recurrent neural networks (RNN) \cite{3zaremba2014learning,4mickey2014neural},  long short-term memory (LSTM) \cite{5zaremba2014learning,6kalchbrenner2015grid}, Neural GPUs \cite{7kaiser2015neural,8price2016extensions,9freivalds2017improving}, and transformers \cite{10transformers,11dehghani2018universal}. These studies were mostly restricted to addition of integers with the same number of digits and did not involve symbolic variables or expressions. The study involving mathematical expressions sought to discover efficient mathematical identities \cite{3zaremba2014learning}. For the studies that considered multiplication, accuracy for the multiplication tasks were either not explicitly reported \cite{5zaremba2014learning,11dehghani2018universal} or involved binary representations \cite{7kaiser2015neural,8price2016extensions,9freivalds2017improving}. 

In this work, we report results for directly evaluating mathematical expressions involving addition, subtraction and multiplication of both positive and negative decimal numbers with variable digits assigned to symbolic variables. The end-to-end process described below does not include any curriculum training or intermediary non-decimal representations.

\section{Experimental Methods}
The training and test data were generated by assigning symbolic variables either positive or negative decimal integers and then describing the algebraic operation to perform. Such expressions were generated as input strings as shown in the example:

\begin{gather*}
    x = 85, y = -523, x \times y \\
    -44455
\end{gather*}

We restrict our variable assignments to the range $x,y \in [-1000,1000)$ and the operations to the set $\{+,-,\times \}$. To ensure that the model embraces symbolic variables, the order in which  and  appears in the expression is randomly chosen. For instance, an input string contrasting from the example shown above might be $y=129,x=531,x-y$. Each input string is accompanied by its target string, which is the evaluation of the mathematical expression. For this study, all targets considered are decimal integers represented at the character level. About 12 million unique samples were thus generated and randomly split into training and test sets at an approximate ratio of 9:1, respectively. 

\begin{table}[t]
\centering
\caption{Example input obfuscation table.}\smallskip
\resizebox{1.0\linewidth}{!}{ 
\begin{tabular}{|c|c|c|c|c|c|c|c|c|c|c|c|c|c|c|}
    \hline
    $x$ & $=$ & $8$ & $5$ & $,$ & $y$ & $=$ & $-$ & $5$ & $2$ & $3$ & $,$ & $x$ & $\times$ & $y$ \\
    \hline
    g & r & f & d & n & p & r & w & d & q & e & n & g & k & p \\
    \hline
\end{tabular}
}
\label{table:1}
\end{table}

\begin{table}[t]
\centering
\caption{Example output obfuscation table.}\smallskip
\resizebox{0.45\linewidth}{!}{ 
\begin{tabular}{|c|c|c|c|c|c|}
    \hline
    $-$ & $4$ & $4$ & $4$ & $5$ & $5$ \\
    \hline
    w & m & m & m & d & d \\
    \hline
\end{tabular}
}
\label{table:2}
\end{table}

The entirety of each input is read and encoded at the character level. The entirety of each output is decoded at the character level. Only after training do the models come to interpret meaning behind the character sequences. One can imagine that a different character mapping be used to obfuscate the meaning assigned by mathematical practice but still be trainable for the models described here to capture the relationships between the individual characters (Table~\ref{table:1}). Mapping such results back to the representations familiar in mathematical practice would yield the same results (Table~\ref{table:2}). 

The input-target pairs were first used to train a self-attentive and feed-forward transformer without recurrence or convolution in a similar manner as the base model previously reported \cite{10transformers}. The self-attention mechanism used is the scaled dot-product attention according to

\begin{equation}\label{eq:1}
    Attention(K,Q,V)=softmax\left(\frac{QK^T}{\sqrt{d}}\right)V,
\end{equation}

\noindent where $d$ is the dimension (number of columns) of the input queries $Q$, keys $K$, and values $V$. By using self-attention, transformers can account for the whole sequence in its entirety and bi-directionally. For multi-head attention with $h$ heads that jointly attend to different representation subspaces at different positions given a sequence of length $m$ and the matrix $H\in\mathbf{R}^{m \times d}$, the result is

\begin{align}
\label{eq:2}
\begin{split}
 MultiHead(H) &= Concat\left(head_1,...,head_h\right)W^O ,
\\
 head_i &= Attention\left(H^W_i,H^K_i,H^V_i\right) ,
\end{split}
\end{align}

\noindent where the projections are learned parameter matrices $H^W_i,H^K_i,H^V_i\in\mathbf{R}^{(d \times d)/h}$ and $W^O\in\mathbf{R}^{(d \times d)}$. 

The same hyperparameters were used as the standard transformer above except for the details that follow. The transformer used in this study is smaller. The encoder consisted of two identical layers, each of which consists of two sub-layers: a multi-head self-attention layer and a fully-connected feed-forward network layer. Layer normalization was used to preprocess the sub-layer inputs. The decoder consisted of two identical layers, each of which consists of three sub-layers: a multi-head self-attention layer, a multi-head attention layer over the output of the encoder stack, and a fully-connected feed-forward network layer. Each multi-headed attention layer consisted of 4 heads. Each fully-connected feed-forward network consisted of 128 neurons. A dropout rate of 0.1 was used to postprocess the output of each sub-layer before it is added to the sub-layer input by the residual connection. The number of training steps was set to 100,000.

\section{Results and Discussion}

The transformer model achieved an accuracy on the test set of 76.1\%. When we analyze the performance by the type of expression, however, we find that the model infers with perfect accuracy symmetric $a(op)a$ expressions such as $x+x$, $y-y$, and $x*x$. Slightly less perfect were asymmetric $a+b$ addition tasks, such as $x+y$ and $y+x$, which had 98\% accuracy. The next challenging tasks involved asymmetric $a-b$ subtraction, such as $x-y$ and $y-x$, which had 49\% accuracy. The model struggled most with asymmetric $a \times b$ multiplication tasks, such as $x \times y$ or $y \times x$, which had only 9\% accuracy. Note that this is a single model trained to perform all the different types of tasks. A summary of the results are shown in Table~\ref{table:3}.

\begin{table}[t]
\centering
\caption{Test performance comparison of inferring mathematical expressions at the character level for different types of expressions for the transformers studied in this work: T - Transformer; UT - Universal Transformer; AUT - Adaptive Universal Transformer.}\smallskip
\resizebox{1.0\linewidth}{!}{ 
\begin{tabular}{c c c c c c c}
    \hline
    \textbf{Type} & $\boldsymbol{a} + \boldsymbol{a}$ & $\boldsymbol{a} - \boldsymbol{a}$ & $\boldsymbol{a} \times \boldsymbol{a}$ & $\boldsymbol{a} + \boldsymbol{b}$ & $\boldsymbol{a} - \boldsymbol{b}$ & $\boldsymbol{a} + \boldsymbol{b}$ \\
    \hline
    \textbf{T} & 1.0 & 1.0 & 1.0 & 0.98 & 0.49 & 0.09 \\
    \hline
    \textbf{UT} & 1.0 & 1.0 & 1.0 & 1.0 & 0.50 & 0.23 \\
    \hline
    \textbf{AUT} & 1.0 & 1.0 & 1.0 & 0.99 & 0.99 & 0.15 \\
    \hline
\end{tabular}
}
\label{table:3}
\end{table}

\begin{figure}[!hbt]
    \centering
    \includegraphics[width=0.85\linewidth]{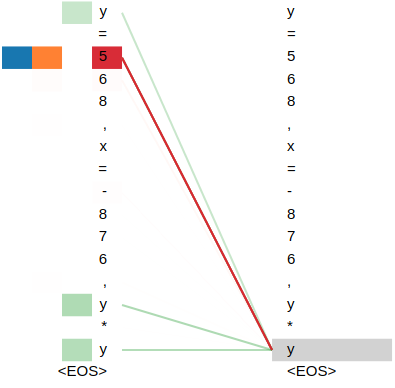}
    \caption{Encoder's self-attention on the last character for one of the transformer layers. The different attention heads are color coded.}
    \label{fig:1}
\end{figure}

\begin{figure}[!hbt]
    \centering
    \includegraphics[width=0.65\linewidth]{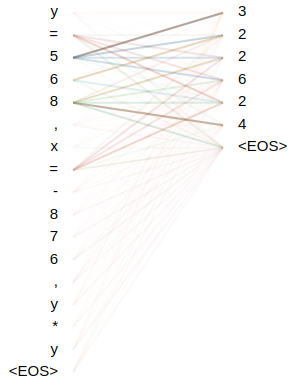}
    \caption{Transformer attention distributions for the decoder's attention on the final layer of the encoder for different decoder layers and attention heads (color-coded). See Fig.~\ref{fig:3} for more visualizations.}
    \label{fig:2}
\end{figure}

The results demonstrate that the transformer can learn to interpret and evaluate symbolic variables and expressions as represented by character strings, performing addition, subtraction and multiplication of both positive and negative decimal numbers. The transformer can correctly utilize the values assigned to symbolic variables for inference. Considering the example input string $y=568,x=-867,y \times y$, the model correctly ignores the value assigned to $x$ and computes $322624$ as the output. The attention visualizations for the encoder's self-attention and decoder's attention on the final layer of the encoder shown in Figs.~\ref{fig:1}~and~~\ref{fig:2}, respectively, illustrate that the output characters attend almost exclusively on the characters representing the assignment to $y$. Furthermore, the order in which $x$ and $y$ assignments occur in the string are handled well, since the accuracy is high despite our data including random variations as mentioned above.

\begin{figure*}[t]
  \begin{subfigure}{6cm}
    \centering\includegraphics[width=5cm]{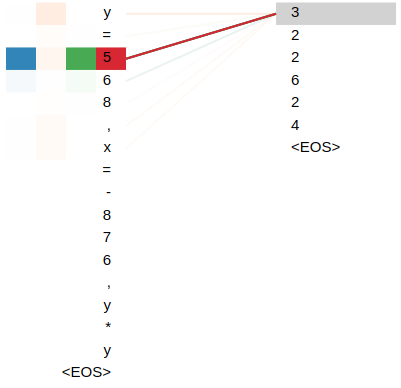}
  \end{subfigure}
  \begin{subfigure}{6cm}
    \centering\includegraphics[width=5cm]{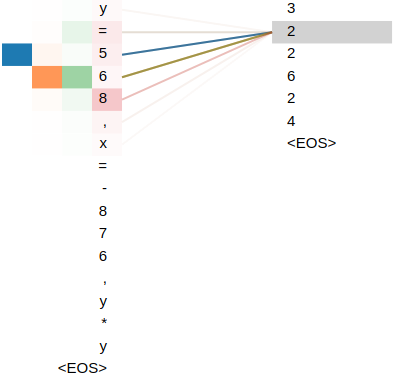}
  \end{subfigure}
  \begin{subfigure}{6cm}
    \centering\includegraphics[width=5cm]{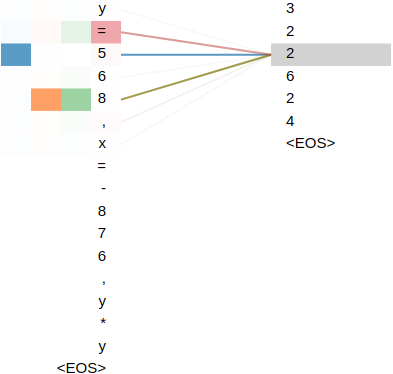}
  \end{subfigure}
  \begin{subfigure}{6cm}
    \centering\includegraphics[width=5cm]{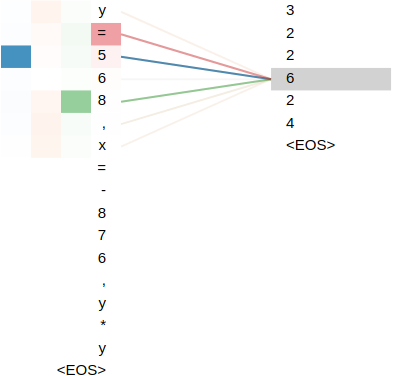}
  \end{subfigure}
  \begin{subfigure}{6cm}
    \centering\includegraphics[width=5cm]{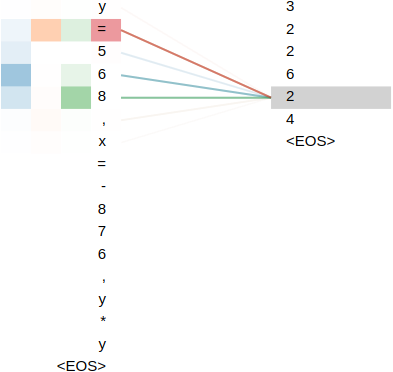}
  \end{subfigure}
  \begin{subfigure}{6cm}
    \centering\includegraphics[width=5cm]{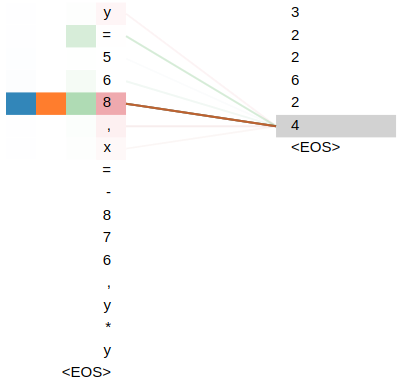}
  \end{subfigure}
  \caption{Extra attention visualizations corresponding to Fig.~\ref{fig:2} for a decoder layer's attention on the final layer of the encoder. The different attention heads are color coded. Attentions are displayed for each output character individually. The first and last characters of the output attend primarily to the first and last characters representing the assignment to $y$, whereas the characters in between do not exhibit such selective attention.}
    \label{fig:3}
\end{figure*}

For the cases inferred incorrectly, the results are very close to the targets. As an example, the value produced for the input sequence $y=-440,x=687,y \times y$ is $-300280$, which is very close to the actual target value of $-302280$ considering the character match accuracy at each position. Only the thousandth place character is incorrect, which is representative of our general observation that one or two of the middle positions are most difficult to correctly infer. Interestingly, the first and last positions of the output attend primarily to the first and last positions representing the assignment to $y$, whereas the output positions in between do not exhibit such selective attention (Fig.~\ref{fig:3}). This confusion could be the reason for the faulty inference of the characters in the middle of the output.

\bigskip

In order to improve the performance for evaluating non-symmetric subtraction and multiplication expressions, the transformer can be augmented with recurrent inductive bias as described by prior work on universal transformers \cite{11dehghani2018universal}. Unlike the standard transformer, the universal transformer can be computationally universal given sufficient memory. At training step $t$, the universal transformer iterates to improve its representations $H^t\in\mathbf{R}^{(m \times d)}$ for the $m$ input positions in parallel with a self-attention mechanism, followed by a recurrent transformation using a depth-wise separable convolution or a position-wise fully-connected layer. The universal transformer was thus reported to achieve state-of-the-art results on translation, natural language understanding, and learning-to-execute tasks similar to this study, outperforming both LSTM RNNs and the standard transformer given the same hyperparameters.

Using the same hyperparameters and dataset described above for the standard transformer, the universal transformer achieved better results on all types of asymmetric $a(op)b$ expressions as shown in Table~\ref{table:3}. The overall accuracy on the tasks is 78.8\%. The most improvement occurred for the $a \times b$ multiplication tasks, which more than doubled in accuracy. It therefore appears that the recurrent inductive bias as implemented in the universal transformer successfully addresses some of the shortcomings of the standard transformer model when using the same hyperparameters.

Since only $a-b$ and $a \times b$ tasks can be improved upon any further, we next add adaptive computation time (ACT) \cite{12graves2016adaptive} to the universal transformer \cite{11dehghani2018universal} in order to devote more processing resources to symbols not interpreted well by the model. For a neural network $R$ with parametric state transition model $S$, output weights $W_y$, output bias $b_y$, input sequence $x_t$, state sequence $s_t$, intermediate update number $n$, intermediate state sequence $s^{n}_t$, intermediate output sequence $y^{n}_t$, augmented input sequence $x^{n}_t$, ACT can be implemented by iterating through each step  of the sequence as follows:

\begin{equation}\label{eq:3}
    x^{n}_t=\left( x_t, \delta_{n,1} \right),
\end{equation}

\begin{equation}\label{eq:4}
s^{n}_t = 
\begin{cases}
S(s_{t-1} , x^{1}_t) & \text{if } n = 1 \\
S(s^{n-1}_t , x^{n}_t) & \text{if } n \neq 1 \\
\end{cases}
\end{equation}

\begin{equation}\label{eq:5}
    y^{n}_t=\left( W_t \cdot s^{n}_t, b_y \right),
\end{equation}

\noindent where $\delta_{n,1}$ is a binary indicator of whether the input step has been incremented at update $n$. An extra sigmoidal halting unit $h$ and its associated weights $W_h$ and bias $b_h$ is added to the network to calculate the halting probability $p^{n}_t$ at intermediate steps up to the total number of updates $N(t)$ according to

\begin{equation}\label{eq:3}
    h^{n}_t= \sigma \left( W_h \cdot s^{n}_t + b_n \right),
\end{equation}

\begin{equation}\label{eq:3}
    N(t) =\min\{n':\sum_{n=1}^{n'} h^n_t >= 1-\epsilon\}
\end{equation}

\begin{equation}\label{eq:4}
    R(t) = 1 - \sum_{n=1}^{N(t)-1} h^n_t
\end{equation}

\begin{align}
    p^n_t = 
    \begin{cases}
        R(t) & \text{if } n = N(t) \\
        h^n_t & \text{if } n \neq N(t)
    \end{cases} ,
\end{align}

\noindent where a small threshold $\epsilon = 0.01$ is used to halt after a single update and an upper bound on updates $N(t) \leq 24$ is imposed. The state and output sequences  and  are calculated as

\begin{equation}
s_t = \sum_{n=1}^{N(t)} p^n_t s^n_t \qquad
y_t = \sum_{n=1}^{N(t)} p^n_t y^n_t .
\end{equation}

As shown in Table~\ref{table:3}, the adaptive universal transformer improves on the $a-b$ tasks almost to perfection but performs much worse on the $a \times b$ tasks, producing an overall higher accuracy of 84.9\%. The adaptive universal transformer may have focused only to improve the $a-b$ tasks because it is more attainable than the $a \times b$ tasks in improving overall efficiency. For 100,000 training steps on a 6-core CPU, the universal transformer requires about 3.5 times the training duration of the vanilla transformer, whereas the adaptive universal transformer requires only 2 times the training duration. This indicates that although improvements can be attributed to more training computational costs, augmenting the universal transformer with ACT improves training efficacy.

\section{Conclusion and Future Work}

The mathematical language understanding demonstrated in this study is foundational for an artificially intelligent agent. The framework and findings discussed should also be transferable to natural language understanding. The symbolic variable assignment is analogous to supporting facts in the bAbi story, question and answering tasks  \cite{13weston2015towards}. Symmetric $a(op)a$ tasks only utilize one of the supporting facts, whereas asymmetric $a(op)b$ tasks utilizes two supporting facts. The symbolic expressions and their evaluation studied here can thus be considered a simplified version of story, question and answering tasks that can be studied and analyzed more expediently and concretely. We expect that future studies will involve more types of symbolic expressions and variables, further elucidating how to improve the shortcomings of existing models to the benefit of more complex natural language understanding problems. More training steps would also improve accuracy. The results described here have been made reproducible with open-sourced software hosted on GitHub \cite{14artitw}.

The transformer model has been shown to work well for a myriad of applications beyond what we typically consider as sequence transduction tasks, e.g. image processing \cite{15parmar2018image}.  More generally, transformers can be applied to problems involving tensors as inputs and tensors as outputs, which is the motivation behind the \textit{Tensor2Tensor} library used in this study \cite{16vaswani2018tensor2tensor}. The attention mechanism of the transformer architecture can be interpreted as a global receptive field that can analyze more than the limited receptive fields, which are often referred to as filters, in convolutional neural networks. We therefore expect that the transformer can serve as a unified model to incorporate and improve upon previous work in churn prediction \cite{17wangperawong2016churn}, information retrieval \cite{18wangperawong2018comparing}, and collaborative filtering \cite{19liu2018collaborative}. The customer's history can be the story or input sequence, and the question can be whether they churn or what item they would choose from recommendations provided.

\bibliographystyle{aaai}
\bibliography{citation.bib}

\end{document}